\documentclass{article}

\usepackage{spconf,amsmath,graphicx}
\usepackage{multirow}
\usepackage{makecell}
\usepackage{balance} 
\usepackage{amsfonts}

\title{Folding Attention: Memory and Power Optimization for \\On-device Transformer-based Streaming Speech Recognition}

\name{%
\begin{tabular}{@{}c@{}}
Yang Li $^*$ \qquad 
Liangzhen Lai $^*$\thanks{$*$ Equal contribution} \qquad 
Yuan Shangguan \qquad
Forrest N. Iandola\\
Zhaoheng Ni \qquad
Ernie Chang \qquad 
Yangyang Shi \qquad 
Vikas Chandra
\end{tabular}}
\address{Meta AI\vspace{-0.6em}}

\usepackage{enumitem}
\ninept
\begin{document}
\maketitle
\begin{abstract}
Transformer-based models excel in speech recognition. Existing efforts to optimize Transformer inference, typically for long-context applications, center on simplifying attention score calculations. However, streaming speech recognition models usually process a limited number of tokens each time, making attention score calculation less of a bottleneck. Instead, the bottleneck lies in the linear projection layers of multi-head attention and feedforward networks, constituting a substantial portion of the model size and contributing significantly to computation, memory, and power usage.

To address this bottleneck, we propose folding attention, a technique targeting these linear layers, significantly reducing model size and improving memory and power efficiency. Experiments on on-device Transformer-based streaming speech recognition models show that folding attention reduces model size (and corresponding memory consumption) by up to 24\% and power consumption by up to 23\%, all without compromising model accuracy or computation overhead.


\end{abstract}
\section{Introduction}
\label{sec:intro}
Transformer-based architectures~\cite{Transformer} have demonstrated notable effectiveness in automatic speech recognition (ASR), spanning various modeling paradigms, including sequence-to-sequence models ~\cite{DongICASSP18, KaritaASRU19, SperberINTERSPEECH18, ZhouINTERSPEECH2018, WangINTERSPEECH20}, neural transducers~\cite{emformer, conformer, ZhangINTERSPEECH20, YehArxiv2019}, Connectionist Temporal Classification~\cite{SalazarICASSP19, ZhangINTERSPEECH20}, and hybrid models~\cite{emformer, PoveyICASSP18, WangICASSP20}. 

The core mechanism of Transformers, known as attention, involves projecting tokens into queries, keys, and values and then comparing queries with keys to calculate attention scores. This attention score calculation exhibits quadratic complexity relative to the number of tokens, which becomes the computation bottleneck for tasks involving long contexts, such as non-streaming full context speech recognition~\cite{conformer,ZhangINTERSPEECH20,YehArxiv2019}. Consequently, many approaches have been focusing on mitigating the complexity of attention score calculations. These methods include exploiting the sparsity~\cite{SparseTransformer, BlockwiseAttention} or low rank~\cite{Linformer} of the attention score matrix or modifying and converting the attention score calculation into a recurrent procedure~\cite{RetNet}.

However, streaming ASR using limited context such as~\cite{dong2019self,emformer,moritz2020streaming} faces a different computation bottleneck --- the linear layers of self-attention and feedforward networks. For low-latency voice assistant or voice search scenarios, streaming ASR models need to process short audio segments within 100 ms at a time, sometimes even downsampling to fewer tokens. With $T$ as the context window size in self-attention and $D$ as the embedding dimension, calculating attention scores has a complexity of $\mathcal{O}(D\times T^2)$, while the linear projection layers of multi-head attention and feedforward networks lead to a complexity of $\mathcal{O}(T\times D^2)$. Due to fewer tokens ($T\ll D$), the former is not a bottleneck anymore; instead, the latter becomes the computation bottleneck.

In addition to the computation overhead, these linear layers serve as the memory and power bottleneck for streaming ASR. Storing their weights requires $\mathcal{O}(D^2)$ memory, significantly exceeding the $\mathcal{O}(H\times T^2)$ consumption for attention scores ($H$ is the number of heads). This heightened memory requirement leads to substantially increased power consumption. While current hardware excels in computation energy efficiency, it demonstrates comparatively lower energy efficiency in memory operations~\cite{MemPowerEfficiency, ComputePowerEfficiency, FactorizedJoiner}. Consequently, these linear layers, responsible for the majority of memory read/write traffic, emerge as the power bottleneck in streaming ASR.

We propose \textit{folding attention} to reduce memory and power consumption in streaming ASR. Folding attention trades minimal attention score computation overhead for significant reductions in memory and power usage in linear layers. Each input token of dimension $D$ is split into $N$ sub-tokens with dimension $\frac{D}{N}$, effectively increasing the token count by $N$. These sub-tokens pass through multi-head attention and a feedforward network, then concatenate into an output token with the original dimension $D$. The weight matrix dimension is reduced by $N$. Compared to standard attention, folding attention offers linear layers $\frac{1}{N}$ computation cost, $\frac{1}{N^2}$ size, and lower memory and power consumption. The increased computation cost for attention score calculation is negligible. Using $N$ folding attention layers to substitute a standard attention layer reduces model size and memory/power overhead substantially without increasing computation. Folding attention applied to multiple Emformer Transducer models shows reductions of 12-24\% in model size (and related memory) and 11-23\% in power consumption on LibriSpeech~\cite{librispeech}, and reductions of 14-23\% in model size (and related memory) and 13-21\% in power consumption on our in-house dataset, all without sacrificing model accuracy or increasing computation overhead.

This paper presents the following contributions:
\begin{itemize}[leftmargin=22pt]
\item We comprehensively analyze the compute, memory, and power overhead in streaming ASR. We identify the bottleneck in the linear layers of self-attention and feedforward networks, rather than in calculating attention scores.

\item On-device applications have strict memory and power budgets. We introduce the technique of folding attention as a means to reduce model size and minimize memory and power consumption in streaming ASR.

\item Through extensive experiments conducted on LibriSpeech and our in-house dataset, we demonstrate the substantial reduction in model size, memory, and power achieved by employing folding attention. This improvement is achieved while maintaining comparable model accuracy and computation cost.
\end{itemize}
\begin{figure}[tb]
\centering
\includegraphics[width=1.0\columnwidth]{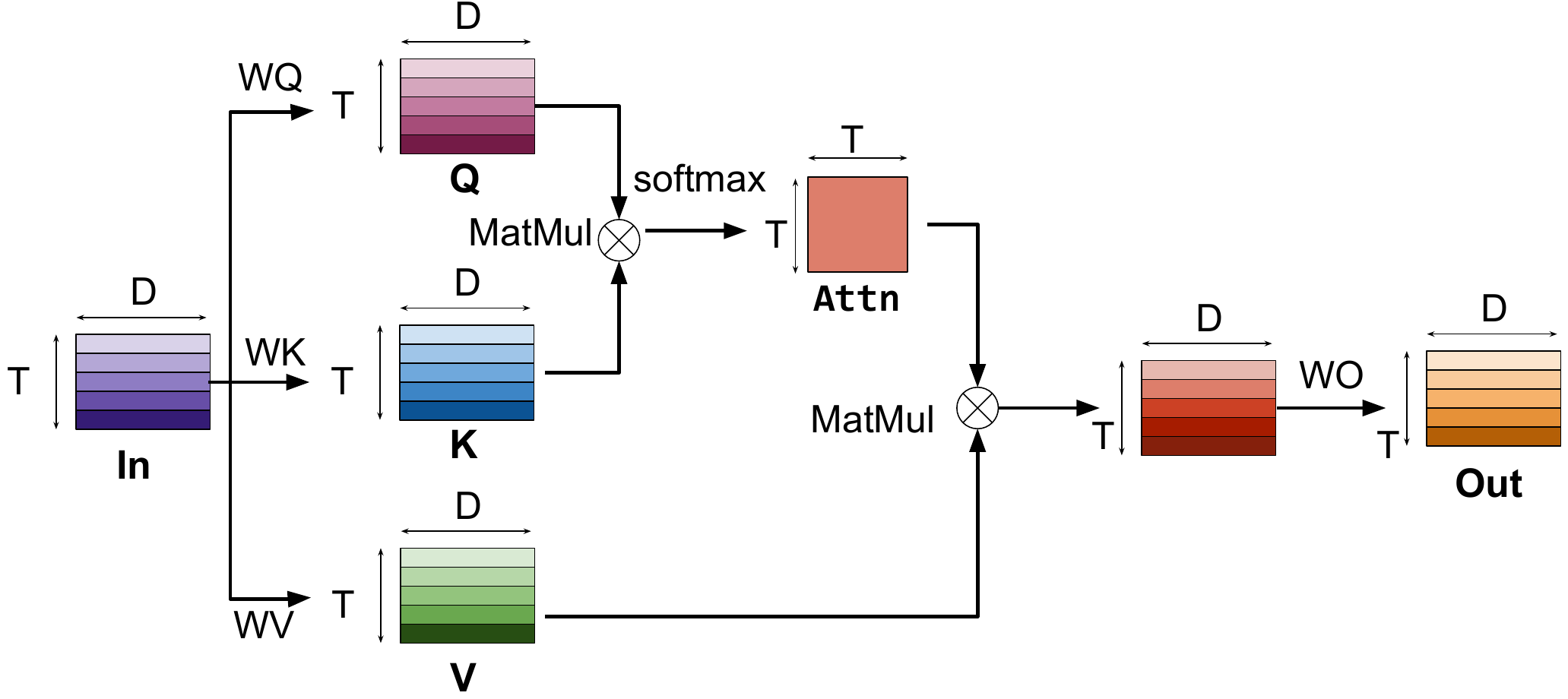}
\caption{Standard self-attention (single-head as an example).}
\label{fig:attention}
\end{figure}

\section{Overhead Analysis of Standard Attention}
In standard self-attention (refer to Figure~\ref{fig:attention}), input tokens are projected into queries, keys, and values using three projection layers: $\mathrm{WQ}$, $\mathrm{WK}$, and $\mathrm{WV}$. Output tokens are obtained through another projection layer, $\mathrm{WO}$. With an embedding dimension of $D$ and a context window size (i.e., token count) of $T$ , each of these projection layers has $D^2$ parameters and has a computational complexity of $\mathcal{O}(T\times D^2)$ and a memory overhead of $\mathcal{O}(D^2)$.

To calculate attention scores, we perform the inner product ($\mathrm{MatMul}$) of each query and key, followed by a $\mathrm{softmax}$ operation to obtain the attention score ($Attn$) for each query-key pair. With $T^2$ pairs of query and key, and the need to store the attention score for each pair under each attention head, the computational complexity for calculating attention scores is $\mathcal{O}(D\times T^2)$ with a memory overhead of $\mathcal{O}(H\times T^2)$ (where $H$ is the number of heads).

As discussed earlier, in streaming ASR, the context length $T$ is much smaller than the embedding dimension $D$. This leads to the computation and memory overhead of linear layers ($\mathcal{O}(T\times D^2)$ and $\mathcal{O}(D^2)$ respectively) significantly surpassing that of attention score calculation ($\mathcal{O}(D\times T^2)$ and $\mathcal{O}(H\times T^2)$). Instead of focusing on reducing the overhead of attention score calculation, as other Transformer optimization techniques~\cite{Linformer, SparseTransformer, BlockwiseAttention, RetNet} do, it becomes more imperative to optimize the linear layers.

\section{Folding Attention}
\label{sec:folding}
Folding attention is designed to decrease the number of parameters in linear layers within self-attention and feedforward networks, thereby mitigating their memory and power consumption.\vspace{0.5em}

\begin{figure}[tb]
\centering
\includegraphics[width=1.0\columnwidth]{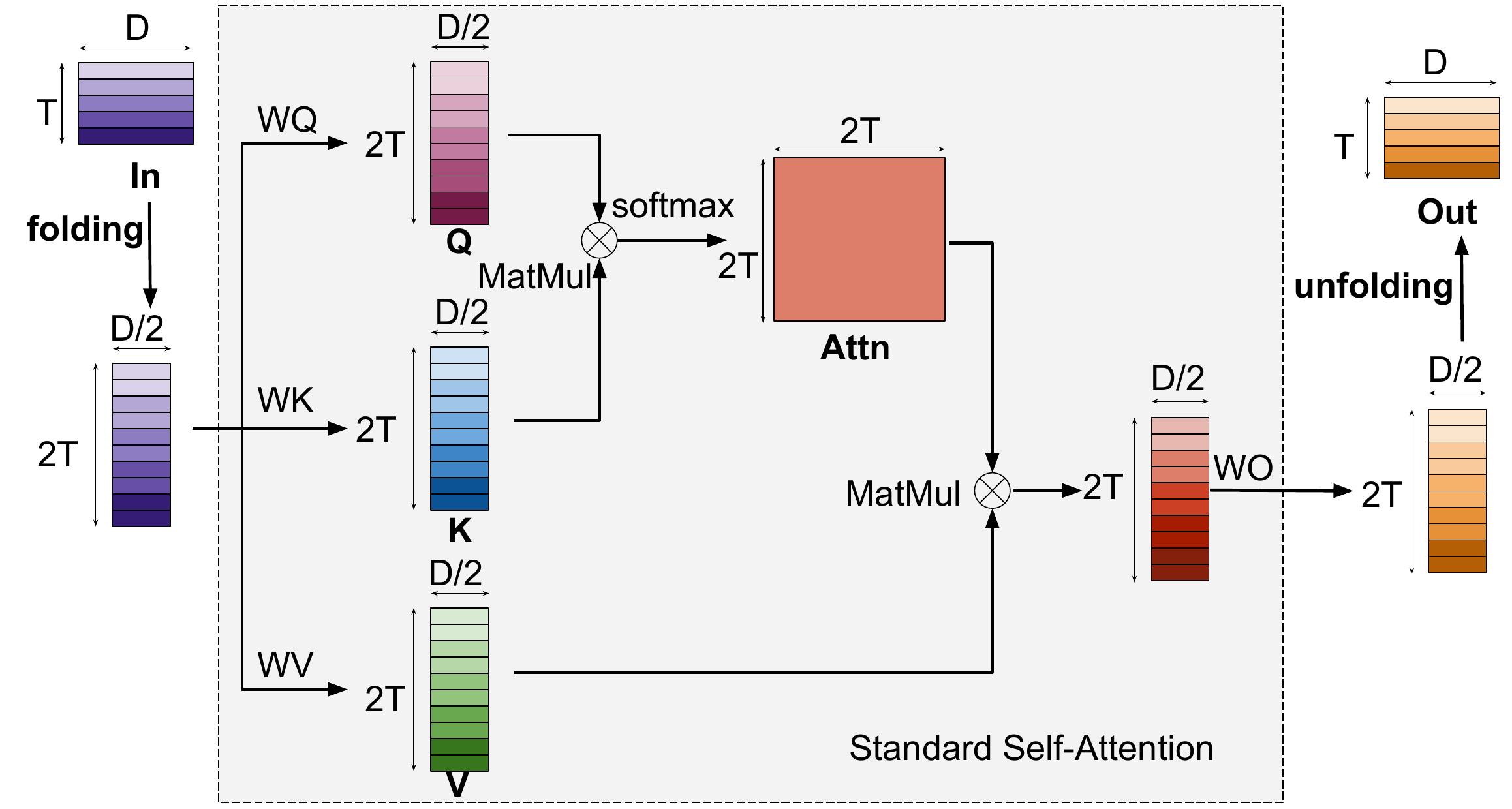}
\caption{Folding attention (folding factor of 2 and single-head as an example). The subsequent feedforward networks (not shown in the diagram) can be repositioned prior to the unfolding operator to achieve a size reduction (fourfold in this example).\vspace{-0.5em}}
\label{fig:folding_attention}
\end{figure}

\noindent\textbf{Design:} The concept of folding attention is depicted in Figure~\ref{fig:folding_attention}. In folding attention, a folding operator and an unfolding operator are introduced respectively before and after the standard self-attention. This folding operator divides an input token into $N$ sub-tokens, where $N$ is the folding factor. This division occurs by assigning the first $\frac{D}{N}$ channels of an original token to the initial sub-token, the subsequent $\frac{D}{N}$ to the second sub-token, and so forth. Consequently, the original $T$ input tokens transform into a sequence of $N\times T$ sub-tokens, each with a dimension of $\frac{D}{N}$. These sub-tokens then proceed through the standard self-attention layer, akin to regular tokens. The standard self-attention layer subsequently yields $N\times T$ new sub-tokens, each with a dimension of $\frac{D}{N}$. Finally, the unfolding operator concatenates every $N$ of them, generating the final $T$ output tokens, each with the original embedding dimension $D$. Folding attention does not change the internal mechanics of self-attention and thus is naturally compatible with any self-attention design or optimization such as multi-head~\cite{Transformer} and multi-query attentions~\cite{MultiQuery}.

\noindent\textbf{Overhead Analysis:} With folding attention, the self-attention layer operates on $N$ times more tokens, yet each token possesses $\frac{1}{N}$ number of channels. This leads to linear layers within the self-attention having $\frac{1}{N}$ computation overhead and $\frac{1}{N^2}$ number of parameters. The attention scores' computation overhead increases by a factor of $N$, and their memory overhead increases by $N^2$ times. However, this increase remains negligible as their original overhead is minimal when the token count is small ($T\ll D$). \textit{This way, folding attention trades a negligible increase in the cost of computing attention scores for a significant reduction in memory and power overhead associated with linear layers.} Overall, for streaming ASR, folding attention reduces attention layers' computation by almost $N$ times and reduces their size and corresponding memory consumption by almost $N^2$ times. Given that the power of streaming ASR is primarily influenced by memory read/write operations~\cite{FactorizedJoiner}, which scale with model size, folding attention also significantly reduces attention layers' power.\vspace{0.5em}


\noindent\textbf{Expressiveness:} Compared to standard self-attention, folding attention maintains an equivalent total number of elements for token embeddings ($N\times T$ token embeddings, each with $\frac{D}{N}$ elements; $T\times D$ total number of elements), ensuring similar expressiveness in this regard. In folding attention, although linear projection layers do not establish dependencies between sub-tokens from the same original token, the inner product of these sub-tokens’ query and key embeddings effectively introduces interdependencies, keeping similar expressiveness as standard self-attention in this aspect. While the linear projection layers in folding attention have fewer parameters and thus reduced expressiveness, incorporating additional folding attention layers can compensate for this. Using $N$ folding attention layers to replace one standard self-attention layer reduces the number of parameters (and related memory) by a factor of $N$ and significantly decreases power consumption while maintaining similar computation overhead. Our experiments demonstrate this can also maintain model accuracy.\vspace{0.5em}

\noindent\textbf{Relation to Prior Work:} FoldedCNN~\cite{kosaian2021boosting} enhances CNN throughput and GPU utilization by unfolding input images, transforming $f$ images with $C$ channels into a single $fC$-channel image. Our work, in contrast, focuses on reducing memory and power in streaming ASR using Transformers. We use a folding operator on input tokens, splitting a $D$-channel token into $N$ sub-tokens with $\frac{D}{N}$ channels each. Similarly, depthwise separable convolution~\cite{chollet2017xception} divides channels into groups, paralleling our method of dividing tokens into sub-tokens.

\section{Evaluation}
\label{sec:results}

\subsection{Datasets}
\label{subsec:datasets}

We conducted experiments on two datasets: LibriSpeech~\cite{librispeech} and an in-house dataset.\vspace{0.5em}

\noindent\textbf{LibriSpeech:} We used 960 hours of its training dataset, extracting 80-dimensional log Mel-filterbank features every 25 milliseconds, with a sliding window of 10 milliseconds. Following the Emformer Transducer work~\cite{emformer}, we used a pre-trained sentence piece model~\cite{kudo2018sentencepiece} to produce 4096-dimensional sentence pieces, plus an additional ``blank'' symbol, as our Transducer models predict. We used \textit{test-clean} and \textit{test-other}, the test sets of Librispeech, for evaluating the model accuracy.\vspace{0.5em}

\noindent\textbf{In-house dataset:} It consists of 23k hours of data sampled from English public videos. The audio was de-identified and aggregated, with personally identifiable information (PII) removed. We distorted the collected audio using simulated reverberation and added randomly sampled additive background noise extracted from publicly available videos. We applied speed perturbations~\cite{ko15_interspeech} to create two additional copies of the training datasets at 0.9 and 1.1 times the original speed. We further applied distortion and additive noise to the speed-perturbed data. This resulted in a total of 127.8k hours of training data. For evaluating the accuracy of models trained on this dataset, we use the following two test sets:
\begin{itemize}[leftmargin=11pt]
\item \textit{dictation:} 5.8k hand-transcribed, de-identified, and aggregated utterances from vendor-collected data where speakers were asked to record unscripted open-domain dictation conversations. The audio was recorded in a variety of noise conditions and speaking volumes.

\begin{table}[t]
    \centering
    \resizebox{1.0\columnwidth}{!}{
    \begin{tabular}{|l|c|c|c|c|c|c|}
    \hline
    model (baseline) & \,A1\, &\,A2\, & \,A3\, & \,A4\, & \,A5\, & \,A6\,\\
    \hline
    \# folding attention layers & 0 & 0 & 0 & 0 & 0 & 0 \\
    \hline
    \# standard attention layers & 6 & 8 & 10 & 12 & 15 & 18 \\
    \hline \hline
    model (folding attention) & B1 & B2 & B3 & B4 & B5 & B6\\
    \hline
    \# folding attention layers & 8 & 8 & 8 & 8 & 10 & 12\\
    \hline
    \# standard attention layers & 2 & 4 & 6 & 8 & 10 & 12\\
    \hline
    \end{tabular}
    }\vspace{-0.5em}
    \caption{Models on LibrsiSpeech: number of folding / standard attention layers in their encoders. Folding factor is 2.}
    \label{table:librispeech_models}
\end{table}

\begin{table}[t]
    \centering
    \resizebox{1.0\columnwidth}{!}{
    \begin{tabular}{|c|c|c|c|c|c|c|}
        \hline
        \multirow{2}{*}{model} & \multirow{2}{*}{\makecell{size \\(M)}} & \multicolumn{2}{|c|}{word error rate (\%)} & \multirow{2}{*}{\makecell{power\\(mW)}} & \multirow{2}{*}{\makecell{compute\\GOPS}} & \multirow{2}{*}{RTF}  \\
         \cline{3-4}
        & & test-clean & test-other & & &  \\
        \hline \hline
        A1  & 33.98 & 5.94 & 13.75 & 27.13 & 2.58 & 0.33 \\
        \hline
        A2 & 40.29 & 5.11 & 12.39 & 31.88 & 2.87 & 0.36 \\
        \hline
        A3 & 46.59 & 4.77 & 11.36 & 36.64 & 3.14 & 0.38 \\
        \hline
        A4 & 52.90 & 4.42 & 11.16 & 41.39 & 3.41 & 0.39 \\
        \hline
        A5 & 62.36 & 4.06 & 10.30 & 48.57 & 3.88 & 0.41 \\
        \hline
        A6 & 71.82 & 4.07 & 10.05 & 55.74 & 4.33 & 0.44 \\
        \hline \hline
        B1 & 27.69& 5.55& 13.25& 22.40& 2.59&  0.39\\
        \hline
        B2 & 34.00& 4.98& 12.21& 27.18& 2.90& 0.42\\
        \hline
        B3 & 40.30& 4.68& 11.05& 31.93& 3.17& 0.43\\
        \hline
        B4 & 46.61& 4.35& 10.57& 36.69& 3.44&  0.43\\
        \hline
        B5 & 54.50& 4.09& 10.03& 42.68& 3.90&  0.45\\
        \hline
        B6 & 62.38& 4.09& 10.10& 48.69& 4.38&  0.50\\
        \hline
    \end{tabular}
    }\vspace{-0.5em}
    \caption{Results on LibrsiSpeech: model size, word error rate, power, compute overhead, and RTF. Models A1--A6 are baseline models; B1--B6 are folding attention models.}
    \label{table:librispeech_results}
\end{table}

\item \textit{messaging:} 13.4k hand-transcribed, de-identified, and aggregated utterances from vendor-collected data where speakers were asked to record audio messages for an unspecific person based on a scripted scenario. These utterances are generally shorter and have a higher signal-to-noise ratio (SNR) than the dictation dataset.
\end{itemize}

\begin{figure}[t]
\centering
\includegraphics[width=1.0\columnwidth]{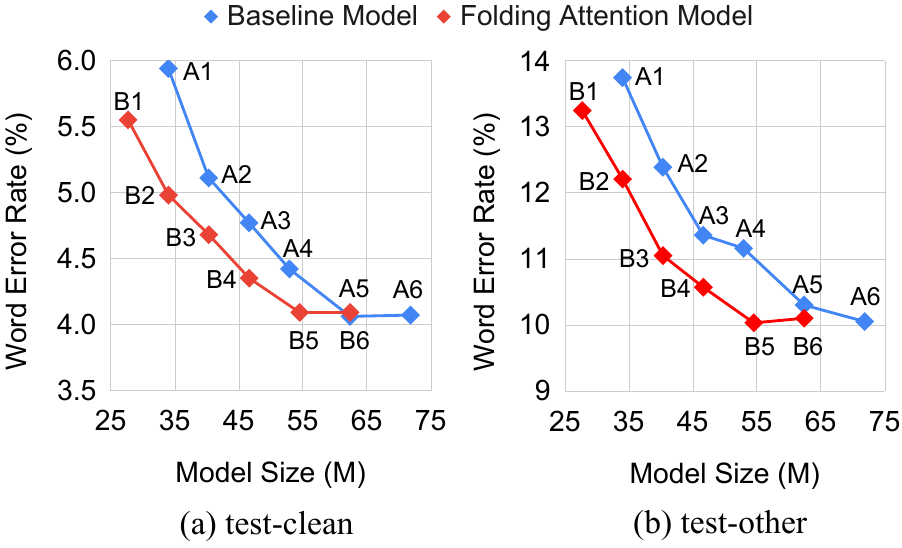}\vspace{-1em}
\caption{Model size vs. word error rate on LibriSpeech.}
\label{fig:librispeech_size}
\end{figure}

\begin{figure}[t]
\centering
\includegraphics[width=1.0\columnwidth]{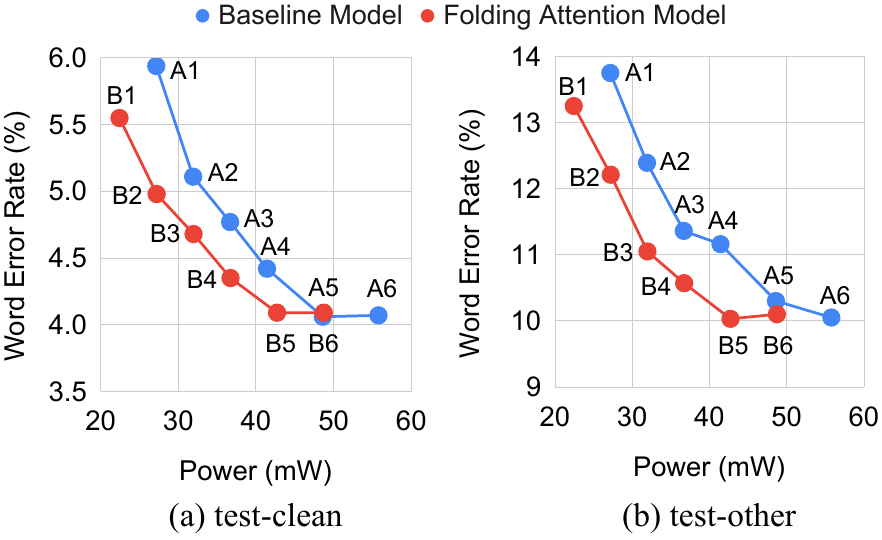}\vspace{-1em}
\caption{Power vs. word error rate on LibriSpeech.}
\label{fig:librispeech_power}
\end{figure}

\subsection{Results on LibriSpeech}
We trained multiple Emformer Transducers~\cite{emformer} on LibriSpeech~\cite{librispeech}, with model details provided in Table~\ref{table:librispeech_models}. Our initial set comprised six baseline models, labeled A1--A6, featuring 6--18 standard attention layers~\footnote{For simplicity, we refer to an attention layer in this section as a combination of a multi-head attention layer followed by a feedforward network.} in their encoders. Additionally, we developed six models, tagged B1--B6, which incorporate 8--12 folding attention layers followed by 2--12 standard attention layers in their encoders. In both baseline and folding attention models, every standard attention layer deploys eight attention heads, and every folding attention layer deploys four attention heads.

We benchmarked these models on a Google Pixel-6 Pro, measuring their Real-Time Factor (RTF) and other critical runtime characteristics. To assess their inference power on a device with a 16 MB cache, we utilized key runtime statistics and state-of-the-art hardware energy efficiency parameters~\cite{MemPowerEfficiency, ComputePowerEfficiency}. The results, including model size, word error rate, power consumption, compute overhead (GOPS),\footnote{GOPS, or Giga Operations Per Second, is the average number of operations from a model that the device must execute per second during streaming speech recognition. It measures the model's computational overhead.} and RTF, are summarized in Table~\ref{table:librispeech_results}.

Figure~\ref{fig:librispeech_size} provides a visual representation of the impact of model size on word error rate. Notably, when comparing models with similar word error rates (B1~vs.~A1, B2~vs.~A2, B3~vs.~A3, B4~vs.~A4, and B5~vs.~A6), folding attention models demonstrate a 12--24\% reduction in model size compared to their baseline counterparts, while maintaining similar computation overhead. The RTF of folding attention models, being well below 1, comfortably meets the streaming ASR requirement.  The marginal increase (0.01–0.06) in their RTF is attributed to additional layers, resulting in a slightly higher interpretive overhead [27]. With an enhanced interpreter, this marginal rise will become even more negligible.

Figure~\ref{fig:librispeech_power} illustrates the relation between power consumption and word error rate. When examining models with comparable accuracy, it becomes evident that folding attention models exhibit an 11–23\% power reduction compared to the baseline models.

\begin{table}[t]
    \centering
    \resizebox{1.0\columnwidth}{!}{
    \begin{tabular}{|l|c|c|c|c|c|c|}
    \hline
    model (baseline) & \,C1\, & \,C2\, & \,C3\, & \,C4\, & \,C5\, & \,C6\,\\
    \hline
    \# folding attention layers & 0 & 0 & 0 & 0 & 0 & 0 \\
    \hline
    \# standard attention layers & 6 & 8 & 10 & 12 & 15 & 18 \\
    \hline \hline
    model (folding attention) & D1 & D2 & D3 & D4 & D5 & D6\\
    \hline
    \# folding attention layers & 8 & 8 & 8 & 8 & 12 & 12\\
    \hline
    \# standard attention layers & 2 & 4 & 6 & 8 & 9 & 12\\
    \hline
    \end{tabular}
    }\vspace{-0.5em}
    \caption{Models on the in-house dataset: number of folding / standard attention layers in their encoders. Folding factor is 2.}
    \label{table:in_house_models}
\end{table}
\begin{table}[t]
    \centering
    \resizebox{1.0\columnwidth}{!}{
    \begin{tabular}{|c|c|c|c|c|c|c|}
        \hline
        \multirow{2}{*}{model} & \multirow{2}{*}{\makecell{size\\ (M)}} & \multicolumn{2}{|c|}{word error rate (\%)} & \multirow{2}{*}{\makecell{power\\(mW)}} & \multirow{2}{*}{\makecell{compute\\GOPS}} & \multirow{2}{*}{RTF}  \\
         \cline{3-4}
        & & dictation & messaging & & &  \\
        \hline \hline
        C1 & 17.20 & 23.13 & 8.08 & \;7.54 & 1.02 & 0.18 \\
        \hline
        C2 & 21.20 & 20.86 & 6.67 & \;9.16 & 1.15 & 0.20 \\
        \hline
        C3 & 25.20 & 19.14 & 5.90 & 10.78 & 1.28 & 0.21 \\
        \hline
        C4 & 29.20 & 18.45 & 5.61 & 12.39 & 1.40 & 0.22 \\
        \hline
        C5 & 35.19 & 17.72 & 5.26 & 14.83 & 1.60 & 0.23 \\
        \hline
        C6 & 41.19 & 17.07 & 4.76 & 17.26 & 1.78 & 0.25 \\
        \hline\hline
        D1 &13.22& 22.44& 7.57& \;5.94& 1.02&  0.21\\
        \hline
        D2 & 17.21& 20.42& 6.57& \;7.56& 1.15& 0.22\\
        \hline
        D3 & 21.21& 18.94& 5.77& \;9.18& 1.27& 0.22\\
        \hline
        D4 & 25.21& 18.17& 5.63& 10.81& 1.40& 0.23\\
        \hline
        D5 & 29.21& 17.63& 5.18& 12.44& 1.60&  0.25\\
        \hline
        D6 & 35.21& 17.04& 4.78& 14.88& 1.80&  0.27\\
        \hline
    \end{tabular}
    }\vspace{-0.5em}
    \caption{Results on the in-house dataset: model size, word error rate, power, compute overhead, and RTF. Models C1--C6 are baseline models; D1--D6 are folding attention models.\vspace{-1em}}
    \label{table:in_house_results}
\end{table}

\begin{figure}[t]
\centering
\includegraphics[width=1.0\columnwidth]{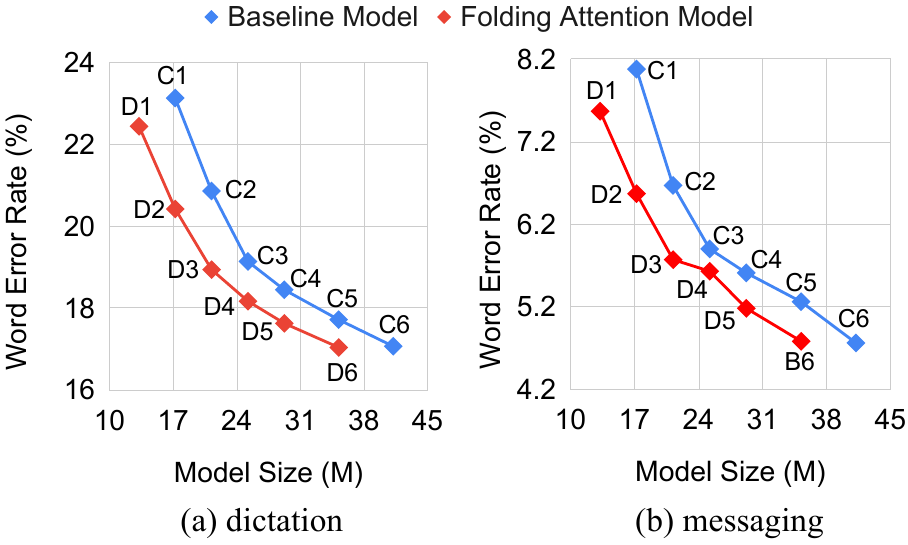}\vspace{-1em}
\caption{Model size vs. word error rate on the in-house dataset.}
\label{fig:in_house_size}
\end{figure}

\begin{figure}[t]
\centering
\includegraphics[width=1.0\columnwidth]{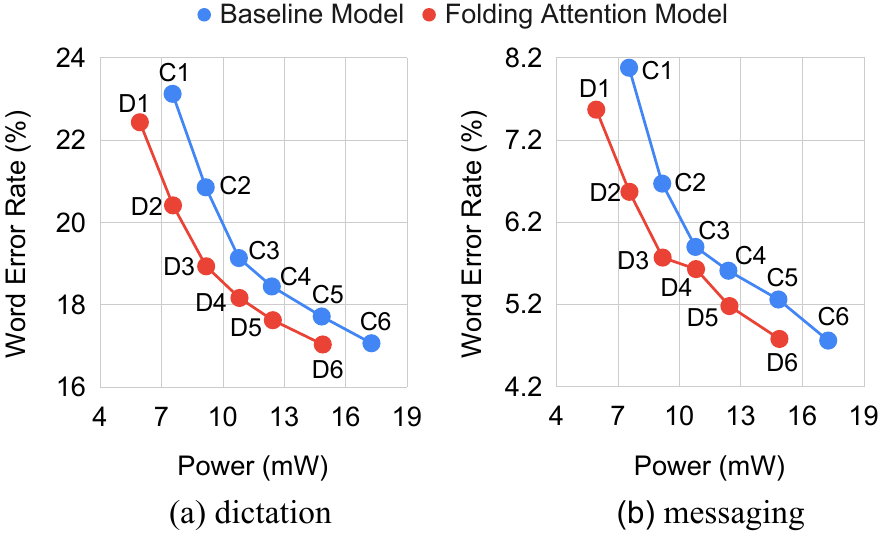}\vspace{-1em}
\caption{Power vs. word error rate on the in-house dataset.}
\label{fig:in_house_power}
\end{figure}

\subsection{Results on the In-House Dataset}
We conducted training on multiple Emformer Transducers using our in-house dataset. Key hyperparameters of these models are summarized in Table~\ref{table:in_house_models}. The baseline models (C1–C6) incorporate between 6 to 18 standard attention layers in their encoders. In contrast, the folding attention models (D1–D6) feature 8 to 12 folding attention layers followed by 2 to 12 standard attention layers in their encoders. In both baseline and folding attention models, every standard attention layer deploys four attention heads, and every folding attention layer deploys two attention heads.

Table~\ref{table:in_house_results} presents a comprehensive overview of these models, detailing their size, word error rate, power, GOPS, and RTF.

Figure~\ref{fig:in_house_size} provides a graphical representation of the relationship between model size and word error rate. Notably, under similar word error rates, folding attention models exhibit a 14–23\% reduction in size compared to their baseline counterparts (comparing models D1~vs.~C1, ..., D6~vs.~C6) while maintaining similar compute GOPS. This reduction in size translates to a noteworthy 13–21\% decrease in power consumption, as illustrated in Figure~\ref{fig:in_house_power}. It is important to note that in ASR inference, power consumption is primarily influenced by memory read/write operations rather than computational operations. With a smaller model size, we can substantially diminish memory read/write operations, consequently reducing power consumption.

\section{Conclusions}
\label{sec:conclusion}
On-device AI applications operate within stringent memory and power constraints, highlighting the critical need for optimization in these domains. We investigate the memory and power challenges associated with Transformer-based streaming ASR. Our analysis reveals a distinctive bottleneck in Transformer-based streaming ASR, predominantly situated within the linear projection layers of self-attention and feedforward networks, as opposed to the attention score calculation, which is the customary focal point of general Transformer optimization strategies.

To address this bottleneck, we introduce the concept of folding attention. The essence of this approach lies in a deliberate trade-off: We accept a negligible increase in computation overhead during attention score calculation in exchange for a noteworthy reduction in memory and power consumption within the linear layers. Upon applying folding attention to on-device streaming ASR, we observed a reduction in model size (and the corresponding memory usage) of up to 24\%, coupled with a decrease in power of up to 23\%, while maintaining similar model accuracy and computation overhead.

These substantial reductions in memory and power consumption are expected to enhance the feasibility of on-device streaming speech recognition, ultimately improving the overall user experience.

\bibliographystyle{IEEEbib}
\bibliography{strings,refs}
\end{document}